\documentclass[11pt,a4paper]{article}
\usepackage{times,latexsym}
\usepackage{url}
\usepackage[T1]{fontenc}

\usepackage[acceptedWithA]{tacl2018v2}

\usepackage{mathtools}
\newcounter{tableeqn}[table]

\newcounter{tablesubeqn}[tableeqn]

\usepackage{relsize}
\usepackage{amsmath, amsthm, amssymb}
\usepackage{mathrsfs}

\usepackage{caption}
\usepackage{subcaption}

\usepackage{enumitem}

\usepackage{color, colortbl}
\PassOptionsToPackage{table,dvipsnames}{xcolor}
\usepackage{multirow,makecell}
\usepackage{rotating}
\usepackage{booktabs}
\usepackage[table]{xcolor}
\usepackage{tabularx}

\usepackage{emoji}
\newcommand{\ku}{\emoji[twitter]{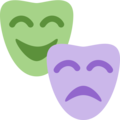}}
\newcommand{\ucambridge}{\emoji[twitter]{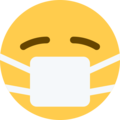}}
\newcommand{\ethz}{\emoji[twitter]{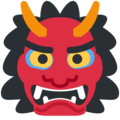}}
\newcommand{\titech}{\emoji[twitter]{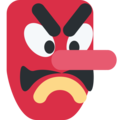}}

\usepackage{todonotes}
\makeatletter
\newcommand*\iftodonotes{\if@todonotes@disabled\expandafter\@secondoftwo\else\expandafter\@firstoftwo\fi}  
\makeatother

\usepackage{euflag}
\usepackage{cleveref}
\crefname{section}{\S}{\S\S}
\crefname{table}{Table}{}
\crefname{figure}{Figure}{}
\crefname{algorithm}{Algorithm}{}
\crefname{equation}{Eq.}{}
\crefname{appendix}{App.}{}
\crefname{prop}{Proposition}{}
\crefname{thm}{Theorem}{}
\crefformat{section}{\S#2#1#3}  

\newcommand{\loss}{\mathscr{L}}

\newcommand{\mathR}{\mathbb{R}}
\newcommand{\mathM}{\mathcal{M}}
\newcommand{\mathMnot}{\backslash\mathcal{M}}
\newcommand{\mathL}{\mathcal{L}}
\newcommand{\mathV}{\mathcal{V}}
\newcommand{\X}{\mathbf{X}}
\newcommand{\Y}{\mathbf{Y}}
\newcommand{\XL}{\mathbf{X}_{\mathcal{L}}}
\newcommand{\XV}{\mathbf{X}_{\mathcal{V}}}

\newcommand{\WQ}{\mathbf{W}^Q}
\newcommand{\WK}{\mathbf{W}^K}
\newcommand{\WV}{\mathbf{W}^V}
\newcommand{\Q}{\mathbf{Q}}
\newcommand{\K}{\mathbf{K}}
\newcommand{\V}{\mathbf{V}}
\newcommand{\OO}{\mathbf{O}}
\newcommand{\scores}{\mathbf{S}}

\newcommand{\M}{\mathbf{M}}

\newcommand{\WO}{\mathbf{W}^O}

\newcommand{\Wone}{\mathbf{W}_1}
\newcommand{\Wtwo}{\mathbf{W}_2}

\newcommand{\relu}{\mathrm{ReLU}}

\newcommand{\ffb}{\mathrm{FFB}}
\newcommand{\mha}{\mathrm{MHA}}
\newcommand{\mab}{\mathrm{MAB}}
\newcommand{\att}{\mathrm{Att}}

\newcommand{\concat}{\mathbin\Vert}

\newcommand{\XM}{\mathbf{X}_{\mathcal{M}}}
\newcommand{\XMnot}{\mathbf{X}_{\backslash\mathcal{M}}}

\newcommand{\cls}{[\texttt{CLS}]}
\newcommand{\sep}{[\texttt{SEP}]}
\newcommand{\img}{[\texttt{IMG}]}
\newcommand{\LN}{\mathrm{LN}}

\newcommand{\vl}{\textsc{V\&L}\xspace}

\newcommand{\bert}{\textsc{BERT}\xspace}
\newcommand{\berts}{\textsc{BERT}s\xspace}
\newcommand{\vilbert}{\textsc{ViLBERT}\xspace}
\newcommand{\vilbertmt}{\textsc{ViLBERT-MT}\xspace}
\newcommand{\lxmert}{\textsc{LXMERT}\xspace}
\newcommand{\visualbert}{\textsc{VisualBERT}\xspace}
\newcommand{\vlbert}{\textsc{VL-BERT}\xspace}

\newcommand{\uniter}{\textsc{UNITER}\xspace}
\newcommand{\ernie}{\textsc{ERNIE-ViL}\xspace}

\newcommand{\volta}{\textsc{Volta}\xspace}

\newcommand{\base}{$_{\textsc{BASE}}$}

\newcommand{\vtheta}{{\boldsymbol \theta}}

\newcommand{\softmax}{\mathrm{softmax}}

\newcommand{\band}{\rowcolor{gray!10}}

\usepackage{xspace,mfirstuc,tabulary}

\newif\iftaclinstructions
\taclinstructionsfalse 
\iftaclinstructions
\renewcommand{\confidential}{}
\renewcommand{\anonsubtext}{(No author info supplied here, for consistency with
TACL-submission anonymization requirements)}
\newcommand{\instr}
\fi

\iftaclpubformat 

\else

\fi

\newcommand\blfootnote[1]{%
  \begingroup
  \renewcommand\thefootnote{}\footnote{#1}%
  \addtocounter{footnote}{-1}%
  \endgroup
}
\title{Multimodal Pretraining Unmasked: A Meta-Analysis and a Unified Framework of Vision-and-Language BERTs}

\author{Emanuele Bugliarello$^{\ku}$~ Ryan Cotterell$^{\ucambridge,\ethz}$~ Naoaki Okazaki$^{\titech}$ Desmond Elliott$^{\ku}$~\\
        $^{\ku}$University of Copenhagen~\;~$^{\ucambridge}$University of Cambridge \\
        $^{\ethz}$ETH Z\"{u}rich~\;~$^{\titech}$Tokyo Institute of Technology \\
        \texttt{emanuele@di.ku.dk},~\;~       \texttt{rcotterell@inf.ethz.ch},~\;~\\ \texttt{okazaki@c.titech.ac.jp},~\;~
        \texttt{de@di.ku.dk}
       }

\date{}

\begin{document}
\maketitle

\begin{abstract}
    Large-scale pretraining and task-specific fine-tuning is now the standard methodology for many tasks in computer vision and natural language processing.
    Recently, a multitude of methods have been proposed for pretraining vision and language \berts to tackle challenges at the intersection of these two key areas of AI.
    These models can be categorised into either single-stream or dual-stream encoders.
    We study the differences between these two categories, and show how they can be unified under a single theoretical framework.
    We then conduct controlled experiments to discern the empirical differences between five \vl \berts.
    Our experiments show that training data and hyperparameters are responsible for most of the differences between the reported results, but they also reveal that the embedding layer plays a crucial role in these massive models.\blfootnote{Pre-print of MIT Press Publication version.}
\end{abstract}

\section{Introduction}\label{sec:intro}
Learning generic multimodal representations from images paired with sentences is a fundamental step towards a single interface for vision-and-language (\vl) tasks. 
In pursuit of this goal, many pretrained \vl models have been proposed in the last year, inspired by the success of pretraining in both computer vision~\citep{sharif2014cnn} and natural language processing~\citep{devlin-etal-2019-bert}. 
All of these \vl  models extend \bert~\citep{devlin-etal-2019-bert} to learn representations grounded in both modalities. They can either be classified as (i) \textit{single-stream}, where images and text are jointly processed by a single encoder~(\emph{e.g.,}~\citealt{Zhou_Palangi_Zhang_Hu_Corso_Gao_2020}), or (ii) \textit{dual-stream}, where the inputs are encoded separately before being jointly modelled~(\emph{e.g.,}~\citealt{tan-bansal-2019-lxmert}). 

The differences in downstream performance between single- and dual-stream models are currently unclear, with some papers claiming the superiority of one family over the other~\citep{lu2019vilbert,chen2020uniter}, while others arguing that it is hard to draw any conclusion~\citep{qi2020imagebert}.

\begin{figure}[t]
   \centering
   \includegraphics[width=\linewidth]{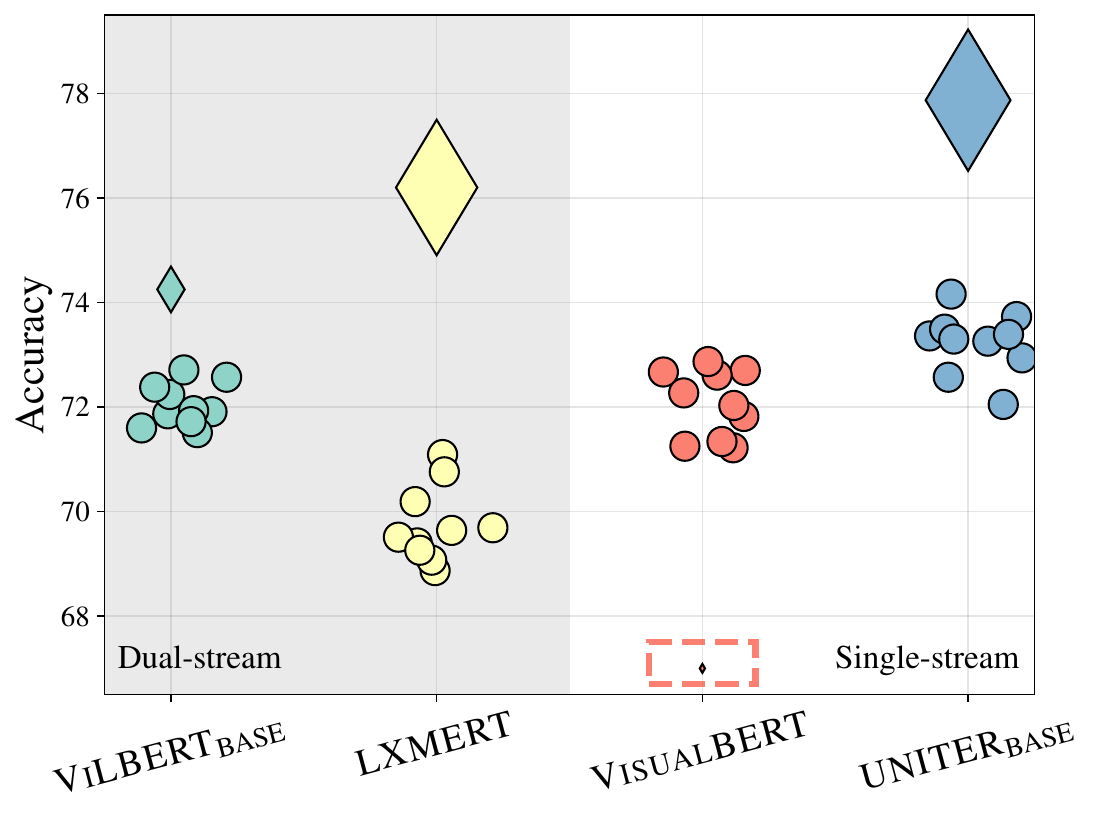}
   \vspace{-0.6cm}
  \caption{How does the amount of pretraining data affect downstream performance of \vl \berts? We find that these models perform \emph{more similarly} when trained in the \emph{same conditions}. This plot shows the results from the papers ($\Diamond$), and when each model is pretrained 10 times on the Conceptual Captions dataset and fine-tuned once on the NLVR2 verification task ($\circ$). The area of a marker is proportional to the amount of pretraining data. The result from the \visualbert paper is highlighted in a dashed box.} \label{fig:intro}
  \vspace{-0.3cm}
\end{figure}

The first goal of this paper is to understand the mathematical differences between single- and dual-stream models. Our analysis leads to a unified framework in which currently proposed architectures, both single- and dual-stream, are particular instances. 
We then implement several of the proposed encoders within this framework to empirically measure their differences in a controlled environment.
We believe this comparative analysis is crucial to better understand and guide future research of massive models in this vibrant area of AI, ensuring progress is not blurred by confounds.

In fact, there are many differences in the protocols used to train \vl \berts.
In order to better understand these models, we conduct a series of controlled studies to investigate whether differences in downstream performance is explained by: (i) the amount of pretraining data and the pretraining objectives (e.g., \cref{fig:comparison}); (ii) the hyperparameters used to control the learning process; (iii) the variance caused by random initialisation when pretraining (e.g., \cref{fig:intro}), (iv) the variance due to fine-tuning multiple times on a downstream task; (v) being single- or dual-stream architectures; or (vi) the choice of the embedding layer.

In summary, our contributions in this paper are:
\begin{itemize}[noitemsep,topsep=1pt]
    \item We introduce a unified mathematical framework in which currently proposed \vl \berts are only a subset of the possibilities.
    \item We release code for \volta (\textbf{V}isi\textbf{ol}inguistic \textbf{T}ransformer \textbf{a}rchitectures),\footnote{\url{https://github.com/e-bug/volta}.} a PyTorch implementation of this framework in order to speed up research in multimodal pretraining.
    \item We conduct a series of controlled studies\footnote{\url{https://github.com/e-bug/mpre-unmasked}.} finding that several models perform similarly when trained under the same conditions.
    \item While we find that single- and dual-stream families perform equally well, performance can differ significantly between two models and the embedding layer plays a key role.
    \item However, these \vl \berts are sensitive to weight initialisation and state-of-the-art claims should not be made from single runs.
\end{itemize}
\section{Vision-and-Language \berts}
Given a sequence of tokens $\{w_1, \dots, w_T \}$ and a set of visual features $\{\mathbf{v}_1, \dots, \mathbf{v}_K \}$, a shared goal of \vl \bert models is to produce cross-modal representations that are useful for downstream tasks grounded in both modalities.

In this section, we first review how these models embed their inputs to the feature space.
Next, we discuss the main differences in the encoders and, finally, highlight a variety of confounds that might affect the performance achieved by these models.

\subsection{Input Embeddings}\label{sec:embs}

\paragraph{Language input}
All \vl \berts adopt the approach of \bert: The input sequence is first tokenized into sub-word units~\citep{wu2016google,sennrich-etal-2016-neural} and two special tokens  $\cls$ and $\sep$ are added to generate the text sequence $\{\cls, w_1, \dots, w_T, \sep\}$.
The embedding of each token is then given by the sum of three learnable vectors, corresponding to its form, position in the sequence and segment~\citep{devlin-etal-2019-bert}.
In addition, \vlbert~\cite{Su2020VL-BERT:} also adds the visual feature of the entire image to each token.

\paragraph{Vision input}
Typically, visual inputs are also very similar across all \vl \berts.
For a given image, a pretrained object detector is used to extract regions of interest (RoIs), representing salient image regions.
For each region, in addition to its feature vector, the object detector also returns the spatial location of its bounding box, which most \vl \berts encode in different ways, analogously to the word position in the language modality.
While most approaches present very similar ways to embed spatial locations, \vlbert relies on a more complex geometry embedding and they are, instead, missing in \visualbert~\cite{li2019visualbert}.
Some models also include a special feature $\img$ that denotes the representation of the entire image (e.g., a mean-pooled visual feature with a spatial encoding corresponding to the full image).
Finally, $\textsc{Pixel-BERT}$~\cite{huang2020pixel} does not rely on an object detector but directly extracts a set of visual embeddings from the raw image.

\subsection{Encoders}\label{sec:encoders}

\paragraph{Single-stream encoders}
The majority of \vl \berts follow the single-stream paradigm~\citep{Su2020VL-BERT:,li2019visualbert,chen2020uniter,Li_Duan_Fang_Gong_Jiang_2020,Zhou_Palangi_Zhang_Hu_Corso_Gao_2020,lin2020interbert,li2020oscar}. 
Here, a standard \bert architecture is given the concatenation of the visual and linguistic features of an image--text pair as input (\cref{fig:trm_blocks}a).
This design allows for an early and unconstrained fusion of cross-modal information.

\paragraph{Dual-stream encoders}
\vilbert~\citep{lu2019vilbert}, \lxmert~\citep{tan-bansal-2019-lxmert}, and \ernie~\citep{yu2020ernie}\footnote{\ernie uses the dual-stream \vilbert encoder.} are based on a dual-stream paradigm.
Here, the visual and linguistic features are first processed by two independent stacks of Transformer layers.\footnote{In practice, \vilbert directly feeds the image representations obtained from the object detector, while \lxmert further processes them through $L_\mathV$ layers.}
The resulting representations are then fed into cross-modal Transformer layers where \emph{intra-modal} interactions are alternated with \emph{inter-modal} interactions (see \cref{fig:trm_blocks}b and c). 
Interestingly, both \vilbert and \lxmert modelled inter-modal interactions in the same way: each stream first computes its query, key, and value matrices, before passing the keys and values to the other modality.
By doing so, these models explicitly constrain interactions between modalities at each layer, inhibiting some of the interactions that are possible in a single-stream encoder while increasing their expressive power by separate sets of learnable parameters.

\subsection{Pretraining Objectives}\label{sec:objectives}

\vl \berts are pretrained by jointly optimising multiple different self-supervised objectives over tokens and image regions through (weighted) scalarisation: $\loss(\vtheta) = \sum_o \lambda_o \loss_o(\vtheta)$.
Here, $\vtheta$ denotes a model's parameters, $\loss_o$ is the $o$-th objective, and $\lambda_o$ is its corresponding weight.
Commonly adopted objectives are of three types: language, vision and cross-modal predictions.

For language prediction, \bert's denoising masked language modelling (MLM) objective is typically used.
MLM replaces some tokens with a $\texttt{[MASK]}$ symbol, which are then predicted by using bidirectional text context and image regions.

The MLM objective has been extended to image regions via masked region modelling objectives.
These typically take the form of either object classification or feature regression, with some papers showing benefits when modelling both~(\emph{e.g.,} \citealt{chen2020uniter}). 
Some models, such as \lxmert, are also optimised over objects' attributes prediction.

Finally, interactions between the two modalities are explicitly enforced by means of cross-modal objectives.
The typical task here is that of image--text matching (ITM; \emph{e.g.,}~\citealt{chen2020uniter}), which extends \bert's next sentence prediction objective to \vl inputs: Given a sequence of tokens and a set of image regions, the model is tasked to predict whether the tokens describe the image.

\subsection{Further Distinctions}\label{sec:distinctions}
So far, we have given an overview of the core components in \vl \berts.
However, there are several implementation differences between them.

For instance, \lxmert presents two main variations to the above description of dual-stream models.
First, in its inter-modal layer, the parameters of the attention sub-layer are shared between the two streams. 
This results in the model learning a single function to contextualise image and text inputs, regardless of which modality plays the role of query or context.
Second, its intra-modal layer only consists of the multi-head attention block. 

Moreover, a wider range of choices can affect the performance of these models.
From the object detector used (and whether it is also fine-tuned during pretraining), to the number of image regions and the maximum text sequence length, to the number of layers and their hidden sizes, to pooling methods and fine-tuning MLP sizes, to the use of text-only data, to optimisation hyperparameters (such as the number of pretraining epochs).

\begin{figure}[t]
\begin{center}
   \includegraphics[width=\linewidth, trim={0cm 0cm 0cm 0cm}, clip]{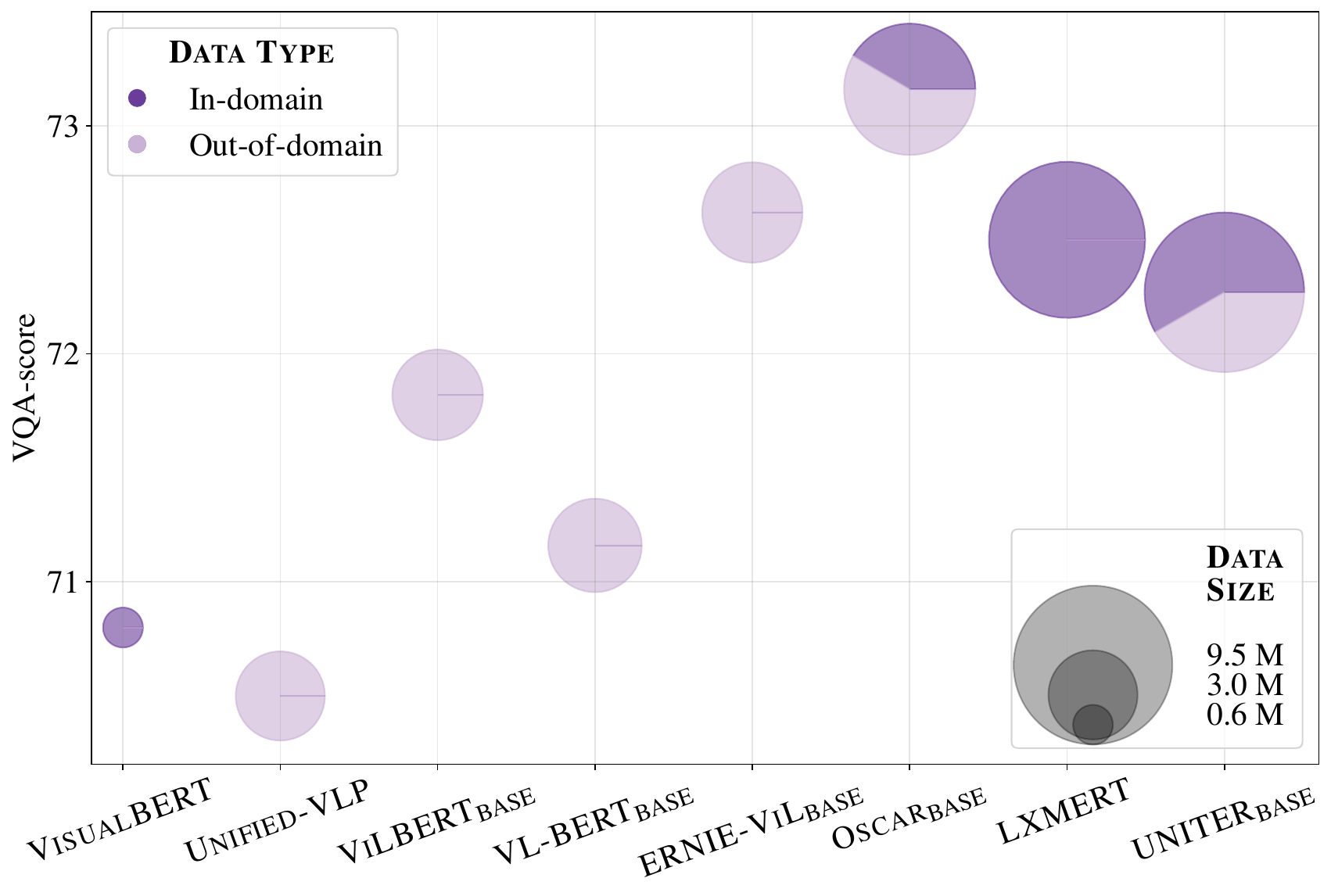}
\end{center}
  \vspace{-0.3cm}
   \caption{Comparison of proposed \vl \berts on VQAv2 (most common downstream task) as a function of their pretraining data (size and type).} \label{fig:comparison}
\end{figure}

Another important distinction is the size and type of pretraining data, which can affect task performance (\cref{fig:comparison}). 
The size of pretraining datasets ranges from $3\text{M}$--$10\text{M}$ image--text pairs, over a range of pretraining tasks. 
The literature distinguishes between ``in-domain'' and ``out-of-domain'' data, each of which may consist of multiple datasets. 
An in-domain dataset overlaps with common downstream tasks, e.g. using VQAv2~\citep{balanced_vqa_v2} as both a pretraining task and a downstream task, while out-of-domain datasets have no expected overlap, e.g. Conceptual Captions~\citep{sharma-etal-2018-conceptual}.

\section{A Unified Framework} \label{sec:framework}

In this section, we unify the recently proposed single-stream and dual-stream architectures under the same mathematical framework. We start by reviewing the Transformer layer, which forms the core of these architectures, then we explain how this layer has been adapted to encode multimodal data in \vl \berts, and introduce a gated bimodal Transformer layer that implements all of the architecture variants as special cases.

\subsection{Transformer Layers}

Transformer-based architectures consist of a stack of Transformer layers~\citep{NIPS2017_7181}, each typically having a multi-head attention block ($\mab$) and a feed-forward block ($\ffb$).

\paragraph{Multi-head attention block}
Given $N_q$ query vectors, each of dimension $d_q$, $\Q\in \mathR^{N_q\times d_q}$, and $N_v$ key--value pairs $\K\in \mathR^{N_v\times d_q}, \V\in \mathR^{N_v\times d_v}$, an attention function $\att(\Q, \K, \V)$ maps queries to output vectors with a scaled dot-product:
\vspace{-2pt}
\begin{equation} \label{eq:att}
    \att(\Q, \K, \V) = \omega\left(\Q\K^\top\right)\V
\end{equation}
\vspace{-15pt}

\noindent where $\omega$ denotes a row-wise, scaled softmax: $\omega_i(\cdot) = \softmax(\cdot/\sqrt{d_q})$. Here, $\scores=\Q\K^\top \in\mathR^{N_q\times N_v}$ is a score matrix that measures the similarity between each pair of query and key vectors.
The output of \cref{eq:att} is a weighted sum of $\V$, in which a value gets higher weight if its corresponding key has a larger dot product with the query.

Multi-head attention (MHA) extends this function by first projecting $\Q, \K, \V$ into $H$ different matrices and computing the attention of each projection  (\cref{eq:att}).
These $H$ different output vectors are concatenated together ($[\concat]$) and the concatenation is projected with a linear transformation $\WO$:
\vspace{-10pt}
\begin{equation} \label{eq:mha}
\begin{array}{c}
    \mha(\Q, \K, \V) = [\OO_1 \mathbin\Vert \dots \mathbin\Vert \OO_H]\WO, \\
    \text{where } \OO_h = \att\left(\Q\WQ_h, \K\WK_h, \Q\WV_h\right).
\end{array}
\end{equation}
Here, $\{\WQ_h, \WK_h, \WV_h \}^H_{h=1}$  and $\WO$ are learned parameters.
Usually, $\mathit{d_q = d_v = d}$, $\WO\in\mathR^{d\times d}$, and $\WQ_h, \WK_h, \WV_h\in\mathR^{d\times d_a}$ where $\mathit{d_a = d / H}$.

Finally, given inputs $\X, \Y \in\mathR^{N\times d}$, a multi-head attention block is defined as:
\vspace{-0pt}
\begin{equation} \label{eq:mab}
    \mab(\X, \Y) = \LN(\X + \mha(\X, \Y, \Y)),
\end{equation}
where $\LN$ is layer normalization~\citep{ba2016layer}.

\paragraph{Feed-forward block}
For an input matrix $\M\in\mathR^{N\times d}$, the feed-forward block is given by: 
\vspace{-3pt}
\begin{equation} \label{eq:ffb}
    \ffb(\M) = \LN(\M + \relu(\M\Wone)\Wtwo),
\end{equation}
where $\Wone,\Wtwo^\top\in\mathR^{d\times d_{\mathit{ff}}}$ are learnable matrices.

\paragraph{Standard Transformer layer}
Let $\X\in \mathR^{N\times d}$ be an embedded input sequence, a standard Transformer layer performing self-attention is a parameterised function $f_\theta: \mathR^{N\times d}\rightarrow \mathR^{N\times d}$ such that:
\vspace{-5pt}
\begin{equation}
    f_\theta(\X) = \ffb(\mab(\X, \X)).
\end{equation}

A stack of $L$ Transformer layers that encodes an input $\X$, such as \bert, is then seen as a sequence of $L$ Transformer layers, each parametrised by $\theta_l$:
\vspace{-5pt}
\begin{equation}
    \text{Encoder}(\X) = f_{\theta_L}\circ\cdots\circ f_{\theta_1}(\X).
\end{equation}

\begin{figure*}[t]
   \centering
   \includegraphics[width=\linewidth, trim={8cm 6cm 8cm 5cm}, clip]{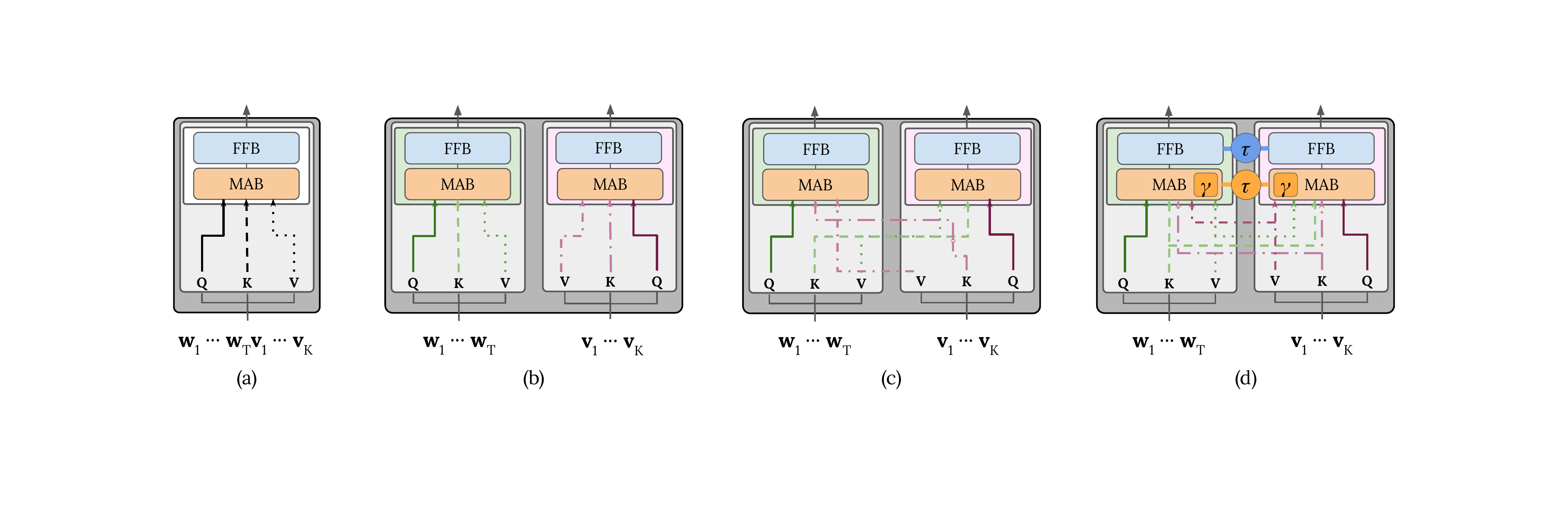}
   \caption{Visualisation of the (a) single-stream, (b) dual-stream intra-modal and (c) dual-stream inter-modal Transformer layers. (d) shows our gated bimodal layer. The inter-modal layer attends across modalities, while the intra-model layer attends within each modality. Ours can attend to either or both.} \label{fig:trm_blocks}
\end{figure*}

\subsection{Single-stream Multimodal Transformers}
Single-stream \vl \berts extend \bert by concatenating the embedded visual inputs $\XV\in \mathR^{N_\mathV\times d}$ and the embedded textual inputs $\XL\in \mathR^{N_\mathL\times d}$ as a single input, hence the name ``single-stream'' (\cref{fig:trm_blocks}a). 
Specifically, $\X = [\XL \concat \XV] \in \mathR^{N\times d}$, where $N = N_\mathL + N_\mathV$, and the attention is over both modalities (\cref{fig:attns}a).
Hence, all single-stream models are of the type defined in the previous section: $\text{Encoder}(\X)$. The various approaches only differ in the initial \vl embeddings, the pretraining tasks, and the training data.

\subsection{Dual-stream Multimodal Transformers}
Both \vilbert and \lxmert concurrently introduced inter-modal and intra-modal layers.

\begin{figure*}[t]
\begin{center}
    \includegraphics[width=\linewidth, trim={0cm 18.5cm 23cm 0cm}, clip]{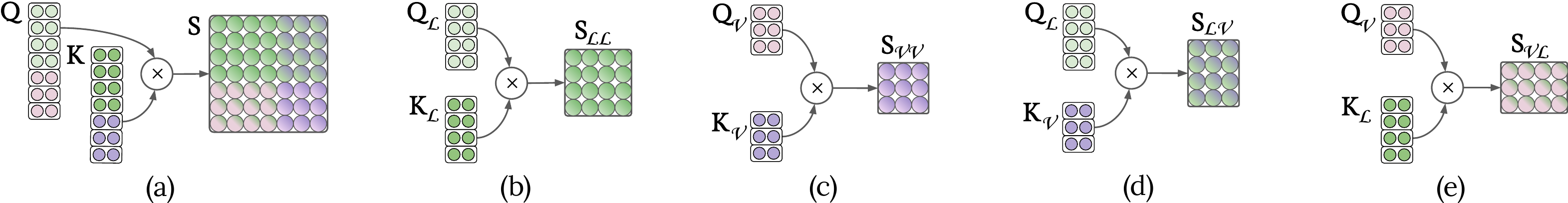}
\end{center}
   \caption{Visualisation of the score matrix for (a) single-stream, (b) text--text, (c) vision--vision, (d) text--vision, and (e) vision--text interactions. Shades of green denote the text modality, while purple ones denote the vision modality. Dual-stream scores are sub-matrices of the single-stream scores matrix.} \label{fig:attns}
\end{figure*}

\paragraph{Inter-modal Transformer layer}
The inter-modal layer explicitly models cross-modal interaction via a cross-modal attention module.
Specifically, let $\mathM \in \{\mathL, \mathV\}$ denote either the linguistic ($\mathL$) or the visual ($\mathV$) modality, and $\mathMnot$ its complementary one.
The inter-modal multi-head attention for modality $\mathM$ is given by (\cref{fig:trm_blocks}c):
\vspace{-5pt}
\begin{equation} \label{eq:inter-mab}
\begin{array}{l}
    \M_{\mathM\mathMnot} = \mab(\XM, \XMnot).
\end{array}
\end{equation}
Note that the second input to the multi-head attention block (\cref{eq:mab}) is taken from the complementary modality, which means the keys $\K$ and values $\V$ in scaled dot-product attention (\cref{eq:att}) operate across modalities (see \cref{fig:attns}d and e). 
The remainder of this layer follows as from \cref{eq:ffb}.

\paragraph{Intra-modal Transformer layer}
The intra-modal layer, on the other hand, is a Transformer layer computing the attention of each modality independently (see \cref{fig:trm_blocks}b). 
For a modality $\mathM$:
\vspace{-5pt}
\begin{equation} \label{eq:intra-mab}
\begin{array}{l}
    \M_{\mathM\mathM} = \mab(\XM, \XM).
\end{array}
\end{equation}
The rest of the layer follows as in \cref{eq:ffb} for \vilbert, while there is no $\ffb$ block in \lxmert.

\subsection{Dual-stream Attentions as Restricted Single-stream Attention}

Recall that in single-stream models the input to a Transformer layer is the concatenation of both modalities, $\X = [\XL \concat \XV]$.
Therefore, in each single-stream attention head, the query representation is given by:
\begin{align}
    \Q = \X\WQ = \begin{pmatrix}
    \XL \\
    \XV
    \end{pmatrix} \WQ
    = \begin{pmatrix}
    \Q_{\mathL} \\
    \Q_{\mathV}
    \end{pmatrix}
\end{align}
where $\begin{pmatrix}
    \cdot_{\mathL} \\
    \cdot_{\mathV}
    \end{pmatrix}$ are the language and visual sub-matrices of the input and the resulting output. A similar expression also holds for the keys $\K$ and values $\V$. We note that the score matrix $\mathbf{S}$ can be defined in terms of four sub-matrices (\cref{fig:attns}a):
\begin{align} \label{eq:score_sub_matrices}
    \scores = \Q \K^{\top} &= \begin{pmatrix}
    \Q_{\mathL} \\
    \Q_{\mathV}
    \end{pmatrix}
    \begin{pmatrix}
    \K_{\mathL}^{\top} & \K_{\mathV}^{\top}
    \end{pmatrix} \nonumber \\
    &= \begin{pmatrix}
    \Q_{\mathL} \K_{\mathL}^{\top} & \Q_{\mathL} \K_{\mathV}^{\top} \\
    \Q_{\mathV} \K_{\mathL}^{\top} & \Q_{\mathV} \K_{\mathV}^{\top}
    \end{pmatrix} \nonumber \\
    &= \begin{pmatrix}
    \scores_{\mathL \mathL} & \scores_{\mathL \mathV} \\
    \scores_{\mathV \mathL} & \scores_{\mathV \mathV}
    \end{pmatrix}
\end{align}

Recall from \cref{eq:att} that the attention matrix is a normalised score matrix $\scores$, so each single-stream layer computes both intra-modal (diagonal of $\scores$) and inter-modal attention (anti-diagonal of $\scores$).
In other words, the dual-stream inter-modal and intra-modal attention functions act as restricted versions of the attention function in any single-stream layer (see \cref{fig:attns}).\footnote{Note that for this to be exact, the learnable parameters of the $\mha$ function need to be shared between modalities (as done, for example, by \lxmert in its inter-modal blocks).}
As a result, by interleaving inter- and intra-modal layers, dual-stream models introduce an \emph{inductive bias} towards which interactions the model enforces in each layer.

\subsection{Gated Bimodal Transformer Layers}

In the previous section, we showed that single-stream attention blocks capture both the inter-modal and intra-modal interactions, separately modelled by dual-stream architectures. We now introduce a general gated bimodal Transformer layer (\cref{fig:trm_blocks}d), in which both single- and dual-stream layers are special cases.
By doing so, we can define existing \vl \berts within a single architecture, which allows us to implement and evaluate several of these models in a controlled environment (see next sections).
In addition to textual $\XL$ and visual embeddings $\XV$, this layer takes a set of fixed binary variables
$\{\boldsymbol{\gamma}, \boldsymbol{\tau}\}$ as part of its input: $\boldsymbol{\gamma}= \{\gamma_{\mathL\mathV}, \gamma_{\mathV\mathL}, \gamma_{\mathL\mathL}, \gamma_{\mathV\mathV}\}$, and $\boldsymbol{\tau}=\{\tau_{\mathit{MHA}}, \tau_{\mathit{LN1}}, \tau_{\mathit{FF}}, \tau_{\mathit{LN2}}\}$.
The $\boldsymbol{\gamma}$ values act as gates that regulate the cross-modal interactions within a layer, while the $\boldsymbol{\tau}$ values control whether the parameters are tied between modalities.

The main difference in our gated layer is in its attention functions, originally defined in \cref{eq:att} and \cref{eq:mha}. 
Here, we extend them to bimodal inputs with controllable multimodal interactions as:
\vspace{-15pt}
\begin{equation} \label{eq:bimha}
\resizebox{.4 \textwidth}{!}{
$
    \mha(\XL, \XV) = [\OO_1 \mathbin\Vert \dots \mathbin\Vert \OO_H]
        \begin{pmatrix}
        \WO_\mathL \\
        \WO_\mathV
        \end{pmatrix}\\
$
}
\end{equation}
where $\WO_\mathL$ and $\WO_\mathV$ are the language and vision output matrices.
The attention output $\att(\Q, \K, \V)$, with a set of gating values $\gamma$ is:
{\small
\begin{align}
  \OO &= \att\left(
    \begin{pmatrix}
        \XL\WQ_\mathL \\
        \XV\WQ_\mathV
    \end{pmatrix},
    \begin{pmatrix}
        \XL\WK_\mathL \\
        \XV\WK_\mathV
    \end{pmatrix},
    \begin{pmatrix}
        \XL\WV_\mathL \\
        \XV\WV_\mathV
    \end{pmatrix}
    ; \boldsymbol{\gamma}\right) \nonumber \\
    &= \att \left(
    \begin{pmatrix}
    \Q_{\mathL} \\
    \Q_{\mathV}
    \end{pmatrix}, 
    \begin{pmatrix}
    \K_{\mathL} \\
    \K_{\mathV}
    \end{pmatrix}, 
    \begin{pmatrix}
    \V_{\mathL} \\
    \V_{\mathV}
    \end{pmatrix}; 
    \boldsymbol{\gamma}\right) \nonumber \\
    &= \omega\left(\scores_{\boldsymbol{\gamma}}\right)\begin{pmatrix}
    \V_{\mathL} \\
    \V_{\mathV}
    \end{pmatrix}
\end{align}
}

Recall from \cref{eq:score_sub_matrices} that the score matrix $\mathbf{S}_\gamma$ can be defined in terms of intra-modal and inter-modal submatrices. 
Here, the gating values $\boldsymbol{\gamma} = \{\gamma_{\mathL\mathL}, \gamma_{\mathL\mathV}, \gamma_{\mathV\mathL}, \gamma_{\mathV\mathV}\}$ define the permitted intra-modal and inter-modal interactions. Let $\varepsilon\rightarrow-\infty$, $\scores_{\boldsymbol{\gamma}}$ is given by:
\begin{align}
    \scores_{\boldsymbol{\gamma}}
    &= \begin{pmatrix}
    \varepsilon^{\gamma_{\mathL\mathL}}\scores_{\mathL \mathL} & \varepsilon^{\gamma_{\mathL\mathV}}\scores_{\mathL \mathV} \\
    \varepsilon^{\gamma_{\mathV\mathL}}\scores_{\mathV \mathL} & \varepsilon^{\gamma_{\mathV\mathV}}\scores_{\mathV \mathV}
    \end{pmatrix}
\end{align}

That is, when an attention gate $\gamma$ is set to $1$, the corresponding sub-matrix tends to $-\infty$, while it is unaltered when $\gamma$ is set to $0$.
By having a sub-matrix that tends to $-\infty$, we can effectively compute the row-wise softmax -- i.e., the attention -- over the other sub-matrix, hence recovering the inter- and intra-modal attentions.\footnote{In practice, our implementation is efficient and does not evaluate sub-matrices whose corresponding gate is set to $1$.}
This is similar to the input masking applied in autoregressive Transformer decoders~\citep{NIPS2017_7181}.

This formulation allows us to control the degree of inter- and intra-modal attention within a layer, allowing us to define existing architectures within a \emph{unified mathematical framework}.
We can recover an inter-modal block (\cref{eq:inter-mab}) by setting $\gamma_{\mathL\mathV} = \gamma_{\mathV\mathL} = 0$ and $\gamma_{\mathL\mathL} = \gamma_{\mathV\mathV} = 1$.
Similarly, the single-stream block (\cref{eq:mab}) can be recovered by setting $\boldsymbol{\gamma} = \mathbf{0}$ and tying the learnable parameters ($\boldsymbol{\tau} = \mathbf{1}$) between the two streams (e.g., $\WQ_\mathL = \WQ_\mathV = \WQ$ in each attention head).

Furthermore, the gated bimodal Transformer layer allows us to model a superset of the few combinations considered thus far for cross-modal fusion by multimodal transformer encoders. 
One may explore asymmetric streams in which the two modalities interact differently with the bimodal inputs, or explore different ways of interleaving conventional single- and dual-stream blocks, or even different levels of parameter sharing.
For example, asymmetric vision-and-language layers might be beneficial for navigation~(\emph{e.g.,} \citealt{hill2021grounded}) or language-conditioned image generation~(\emph{e.g.,} \citealt{cho-etal-2020-x}).
An exploration of these possibilities is left for future work.

\section{Experimental Setup}

In this section, we present the experimental setup for our controlled studies on \vl encoders.

\paragraph{\volta}
In order to facilitate research and development of \vl pretraining, we release \volta (\textbf{V}isi\textbf{ol}inguistic \textbf{T}ransformer \textbf{a}rchitectures), an implementation of our unified framework in PyTorch~\citep{NEURIPS2019_9015}.
Our code is built on top of the \vilbertmt repository,\footnote{\url{https://github.com/facebookresearch/vilbert-multi-task/}.} based on PyTorch-Transformers, due to its support to a wide range of \vl tasks.
We stress that it is important, for this study, to have a unified implementation that allows us to remove possible confounds due to implementation details and effectively measure differences given by the proposed architectures.

\paragraph{Implementation details} 
\vl \berts typically extract image features using a Faster R-CNN~\citep{NIPS2015_5638} trained on the Visual Genome dataset~(VG;~\citealt{krishna2017visual}), either with a ResNet-$101$~\citep{he2016deep} or a ResNeXT-$152$ backbone~\citep{xie2017aggregated}.
The number of features varies from $10$ to $100$. 
Our models are trained with $36$ regions of interest extracted by a Faster R-CNN with a ResNet-$101$ backbone~\citep{DBLP:conf/cvpr/00010BT0GZ18}. 
Each model is initialized with the parameters of \bert, following the approaches described in the original papers.\footnote{Only \citet{tan-bansal-2019-lxmert} reported slightly better performance when pretraining from scratch but they relied on large corpora of in-domain, human-annotated data.}
Randomly initialized weights are initialized following the standard approach in PyTorch-Transformers (on which these models built on): fully-connected and embedding layers are initialized from a normal distribution with mean $0.0$ and standard deviation $0.02$, bias vectors are initially set to $0.0$, and the Layer Normalization weight vector to $1.0$.
We train all models on $4$ NVIDIA P$100$ GPUs and rely on gradient accumulation to obtain larger batches when needed.
The parameter sets giving the best validation performance based on the pretraining objective are used for downstream tasks.

\begin{table}[t]
\setlength\tabcolsep{3pt}
\footnotesize
\renewcommand{\arraystretch}{1.25}
\center
  \begin{tabular}{llrrc}
  \toprule
  \textbf{Dataset} & \textbf{Image Source} & \textbf{Train} & \textbf{Test} &  \textbf{Metric} \\
  \toprule
      VQAv2 & COCO & 655K & 448K & VQA-score \\
      GQA & COCO+Flickr & 1.1M & 12.6K & Accuracy \\
  \midrule
      RefCOCO+ & COCO & 120K & 10.6K & Accuracy \\
      RefCOCOg & COCO & 80K & 9.6K & Accuracy \\
  \midrule
    NLVR2 & Web Crawled & 86K & 7K & Accuracy \\
    SNLI-VE & Flickr & 529K & 17.9K & Accuracy \\
  \midrule
    COCO & COCO & 567K & 1K & Recall@1 \\
    Flirckr30k & Flickr & 145K & 1K & Recall@1 \\
  \bottomrule
  \end{tabular}
  \caption{Statistics of the downstream \vl tasks.}\label{tab:tasks}
\end{table}

\paragraph{Pretraining}
As discussed in \cref{sec:distinctions}, \vl \berts have been pretrained on datasets of varying size and type.\footnote{\vlbert also adds text-only data to avoid overfitting on short and simple sentences typical of \vl datasets.}
In this paper, we pretrain all of our models on the Conceptual Captions dataset (CC;~\citealt{sharma-etal-2018-conceptual}), which consists of $3.3\text{M}$ images with weakly-associated captions automatically collected from billions of web pages. This stands in contrast to other datasets, e.g. COCO~\citep{10.1007/978-3-319-10602-1_48} or VQA~\citep{VQA}, where the images are strongly-associated with crowdsourced captions or question--answer pairs.
The Conceptual Captions dataset is a good candidate for learning generic multimodal representations because of its size, that it was scraped from the Web, and that it has a broad coverage of subject matter.\footnote{We also expect this type of dataset will be easier to collect for low-resource languages in the future.}
Note that due to broken links, and a subsequent pruning phase, where images also found in the test sets of common \vl tasks\footnote{The datasets listed in \cref{tab:tasks}, Visual 7W~\citep{7780909}, RefCOCO~\citep{kazemzadeh-etal-2014-referitgame}, GuessWhat~\citep{Vries_2017_CVPR} and VCR~\citep{zellers2019vcr}.} are removed, we pretrain all our models on $2.77\text{M}$ image--caption pairs from Conceptual Captions.

\paragraph{Downstream evaluation tasks}
We consider the most common tasks used to evaluate \vl \berts, spanning four groups: vocab-based VQA~\citep{balanced_vqa_v2,Hudson_2019_CVPR}, image--text retrieval~\citep{10.1007/978-3-319-10602-1_48,10.1109/ICCV.2015.303}, referring expression~\citep{kazemzadeh-etal-2014-referitgame,7780378} and multimodal verification ~\citep{suhr-etal-2019-corpus,xie2019visual}.
See \cref{tab:tasks} for details.\footnote{Following previous work, accuracy in referring expression is evaluated on the region proposals of \citet{Yu_2018_CVPR}.}
For each model, the parameter set giving the best performance in the validation set was used for test.

\section{Results}

We perform carefully controlled experiments to investigate the possible reasons for the reported difference in performance between \vl \berts.

\subsection{Unified Data and Reimplementation}\label{sec:unified_data}

We start by examining the performance of \vl \berts pretrained on the same $2.7$M CC dataset. Recall from \cref{fig:comparison} that \vl \berts have been pretrained on different combinations of datasets, which may explain most of the claimed differences in downstream task performance. 
Here, we evaluate three models with official released code: \vilbert,\footnote{\vilbert was trained as described in \citet{Lu_2020_CVPR}.} \lxmert and \vlbert. 

\begin{figure}
\begin{center}
  \includegraphics[width=\linewidth, trim={0cm 0cm 0cm 0cm}, clip]{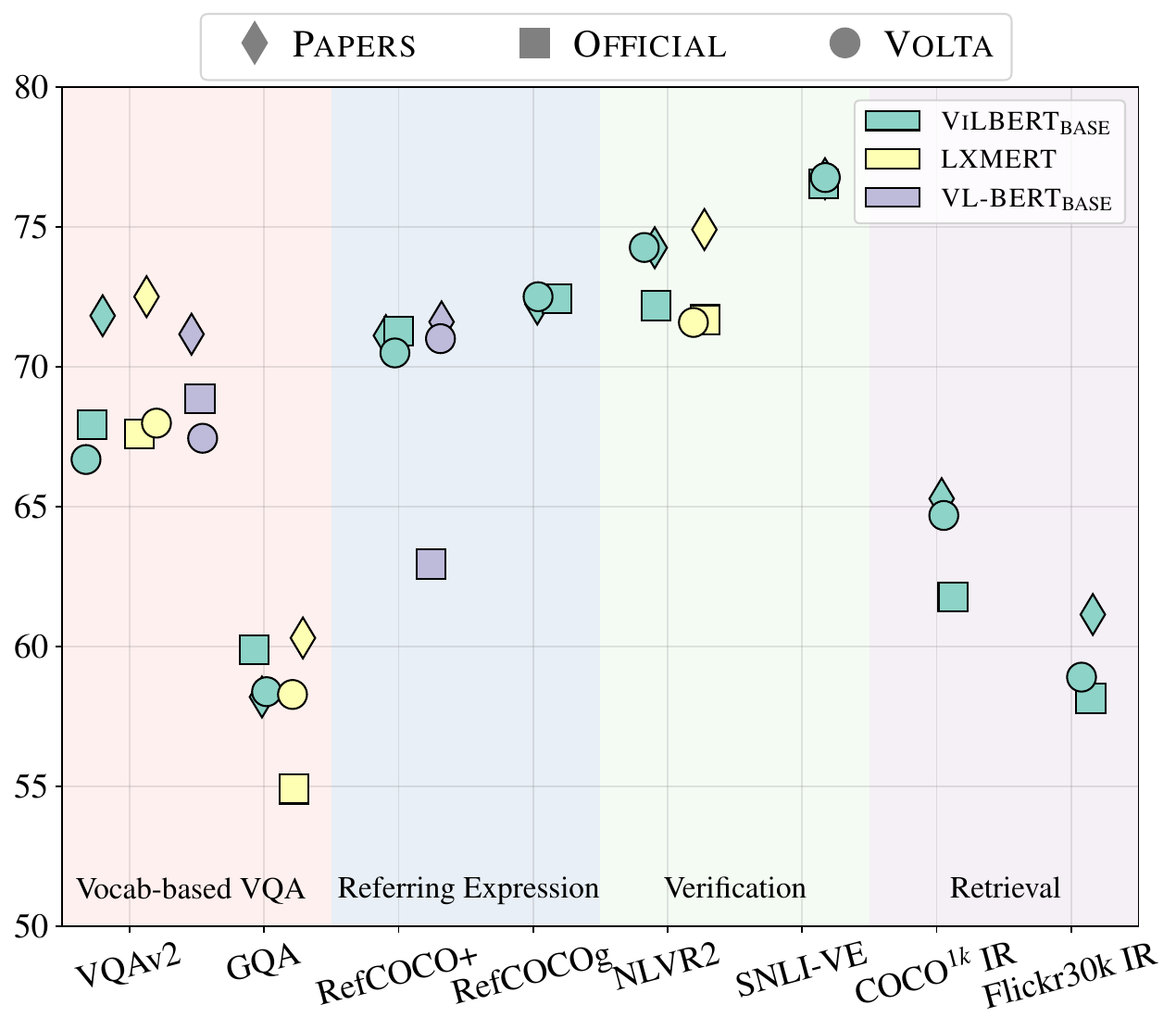}
\end{center}
  \vspace{-0.3cm}
  \caption{Unified data and reimplementation results. Performance of selected \vl \berts on multiple tasks from the original papers ($\Diamond$), and when pretrained on $2.7\text{M}$ Conceptual Captions with their official code ($\square$) or in \volta ($\circ$).} \label{fig:reproduction}
\end{figure}

\paragraph{Same data, similar performance}
\cref{fig:reproduction} shows the results of controlling the pretraining data and pretraining tasks. The results from the papers are reported ($\Diamond$), alongside our training of these models using the official code ($\square$).
There is a drop in performance for the models we trained on the VQAv2, NLVR2, and image retrieval tasks, compared to the performance reported in the papers.
This is not surprising given that the $\square$ models were pretrained on less data than the papers.
In particular, given that \vilbert was also pretrained on CC but with more image--text pairs, our results corroborate previous studies showing diminishing returns with pretraining data size (\emph{e.g.,} \citealt{lu2019vilbert,Li_Duan_Fang_Gong_Jiang_2020}).
However, the claimed performance gaps between these models \emph{narrows} when they are pretrained on the same data.
For instance, according to the literature, \lxmert was clearly the best model in VQA tasks, which is likely due to its use of large, in-domain data and a VQA pretraining objective.\footnote{Surprisingly, for VQAv2, each of these models used different proportions of the validation set during training. 
In our experiments, instead, we use the official training set, which explains why the largest drops in performance are seen here.}

\paragraph{\volta implementation} We also implemented these models in \volta and trained them using their official procedures and hyperparameters.
\cref{fig:reproduction} shows that the performance of each of these models ($\circ$) closely follows the official implementations in these downstream tasks, confirming the correctness of our framework.
There are, however, some larger differences for some of the tasks: in VQAv2, we now see that \vilbert performs slightly worse than the other models in (contrarily to what we obtained with the official code), and in GQA, \lxmert closes the gap with \vilbert.
\vilbert's performance on NLVR2 and COCO image retrieval increases by $2$--$3$ points in the \volta framework.
As \volta is based on the \vilbert code base, these differences might be due to weight initialisation, an hypothesis that we test in later sections.

\paragraph{}
With this first study, we have seen that the performance of these \vl \berts is similar when they are trained on the same data.
Moreover, we demonstrated the correctness of our implementations in \volta, in which these models are built following the unified framework introduced in \cref{sec:framework}.
Nevertheless, there are still many possible confounds in the training procedures adopted by these models that might interfere with a fair comparison of these architectures.
In the next section, we control these variables to unmask the true gains introduced by a number of multimodal encoders.

\subsection{Controlled Setup}
We define a fixed set of hyperparameters to evaluate \vilbert, \lxmert, \vlbert, \visualbert, and \uniter on four downstream tasks: VQAv2, RefCOCO+, NLVR2 and Flickr30K.
\begin{itemize}[noitemsep,topsep=1pt]
    \item \textbf{Inputs:} Each model used a different maximum number of tokens and \lxmert did not have an overall \texttt{[IMG]} feature. We fix the same maximum number of tokens and add the \texttt{[IMG]} feature to each architecture.
    \item \textbf{Encoders:} We noticed that \vilbert used higher dimensional representations for the visual stream. We fix the same dimension as in the linguistic stream for a comparison that is fairer comparison against \lxmert, and more intuitive with the single-stream models. 
    \item \textbf{Pooling:} While \vlbert is the only architecture that does not have a pooling layer, other \vl \berts use it for the image--text matching objective. We fix the models to use use multiplicative pooling~\citep{lu2019vilbert} for all the models in order to separately learn sentence-level and image-level representations and also model their interactions.
    \item \textbf{Pretraining objectives:} Each model uses a different set of pretraining objectives. We fix them to three: MLM, masked object classification with KL-divergence,\footnote{\citet{chen2020uniter} showed that this object classification objective is the single best one for masked regions prediction.} and ITM.
    \item \textbf{Fine-tuning:} We fine-tune each model using the same protocols and sizes for the MLPs.
    \item \textbf{Hyperparameters}: While \vilbert and \vlbert were originally pretrained for $10$ epochs, \lxmert was pretrained for $20$. We fix the number of pretraining epochs to $10$, and set other hyperparameters, e.g., learning rate or its warm-up proportion, to a set of values from the original papers that led to smooth training of all the models, with training curves that closely followed the ones obtained with the original hyperparameters.\footnote{Configuration files of this setup are part of our repository.}
\end{itemize}

\begin{table}
\setlength\tabcolsep{3pt}
\center
  \resizebox{0.48\textwidth}{!}{
  \begin{tabular}{l  c@{\hspace{2\tabcolsep}}  c  c@{\hspace{2\tabcolsep}}  ccc  c@{\hspace{2\tabcolsep}}  cc}\\
  \toprule
  \multirow{2}{*}{\small \shortstack{\textbf{Model}}} && VQAv2 && RefCOCO+ && NLVR2 && \multicolumn{2}{c}{\small Flickr30k} \\
  \cmidrule(r){3-3}
  \cmidrule(r){5-5}
  \cmidrule(r){7-7}
  \cmidrule(r){9-10}
  && test-dev && test$^d$ && test-P && test IR & test TR \\
  \midrule
  \band ViLBERT\base  && 68.7 && 71.4 && 72.4 && 59.8 & 76.7 \\
  LXMERT             && 67.1 && 68.8 && 69.1 && 50.4 & 62.5 \\
  \band VL-BERT\base  && 68.3 && 71.1 && 72.6 && 57.9 & 68.5 \\
  VisualBERT         && 68.2 && 69.7 && 71.3 && 61.1 & 75.5 \\
  \band UNITER\base   && 68.8 && 71.9 && 72.9 && 60.9 & 74.2 \\
  \bottomrule
  \end{tabular}
  }
  \caption{Results with our controlled setup. Each model is pretrained using the \volta framework with the same fixed hyperparameters on the $2.7$M CC dataset, and fine-tuned on downstream tasks. \label{tab:controlled}}
\end{table}

\begin{figure*}
\centering
\begin{subfigure}{.25\textwidth}
  \centering
  \includegraphics[width=\linewidth]{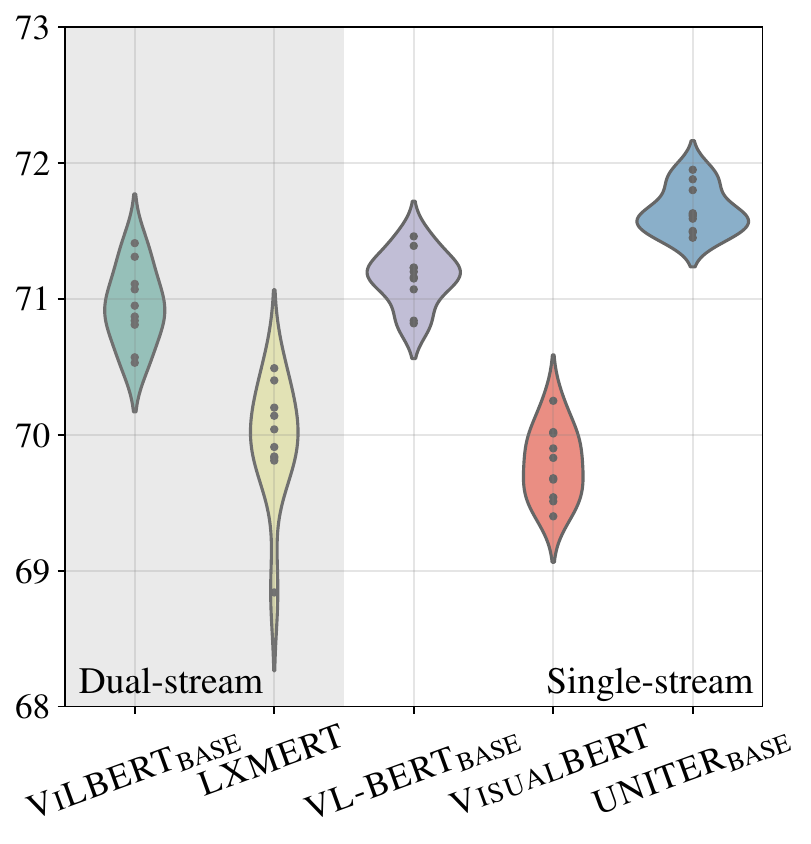}
  \caption{RefCOCO+}
  \label{fig:refcoco_pre}
\end{subfigure}%
\begin{subfigure}{.25\textwidth}
  \centering
  \includegraphics[width=\linewidth]{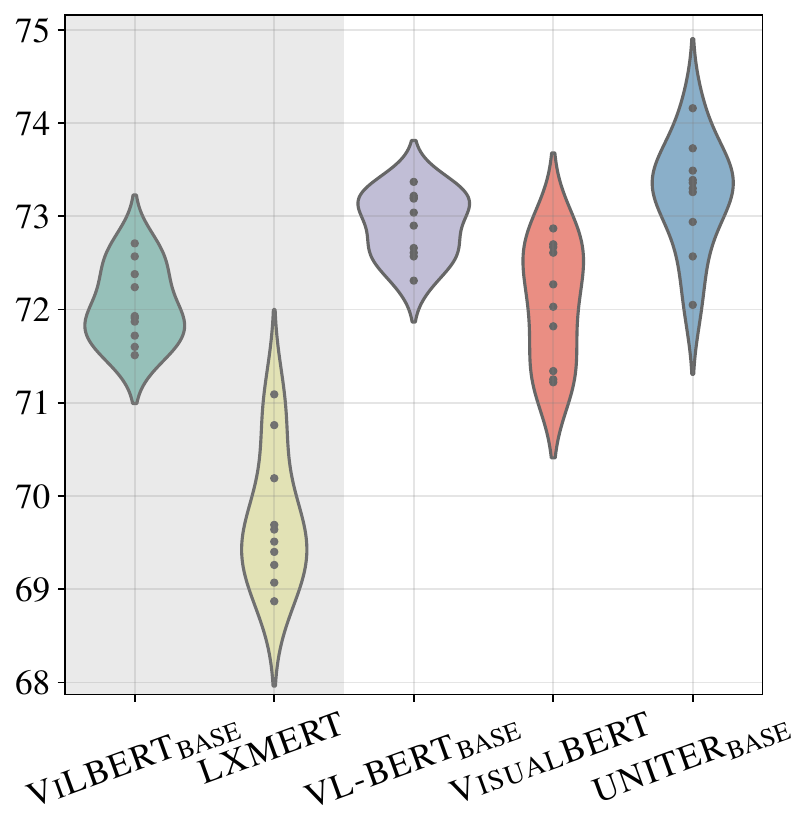}
  \caption{NLVR2}
  \label{fig:nlvr_pre}
\end{subfigure}%
\begin{subfigure}{.25\textwidth}
  \centering
  \includegraphics[width=\linewidth]{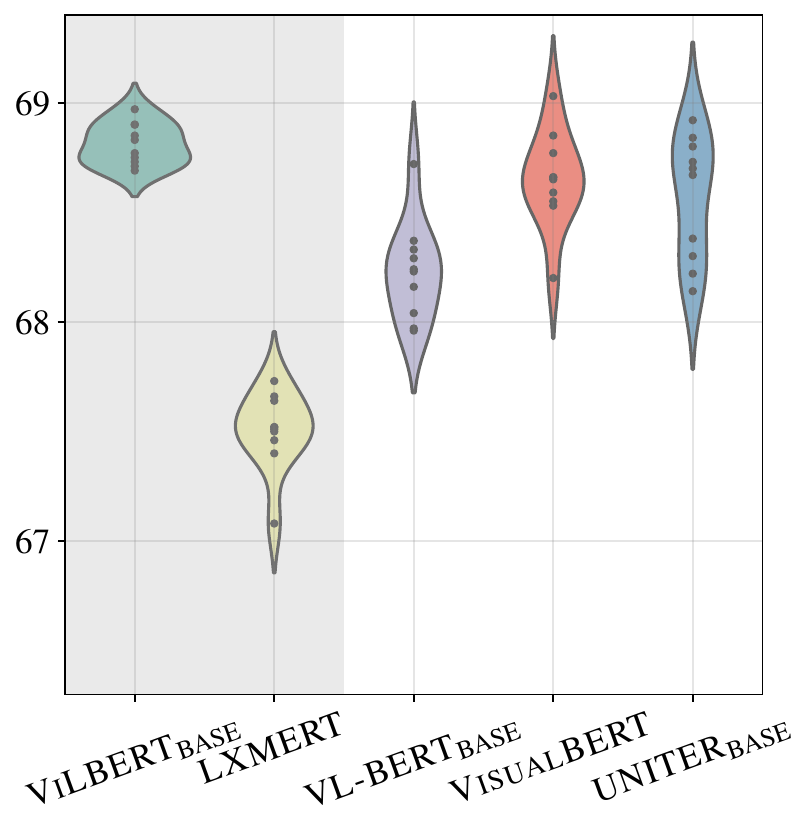}
  \caption{VQAv2}
  \label{fig:vqa_pre}
\end{subfigure}%
\begin{subfigure}{.25\textwidth}
  \centering
  \includegraphics[width=\linewidth]{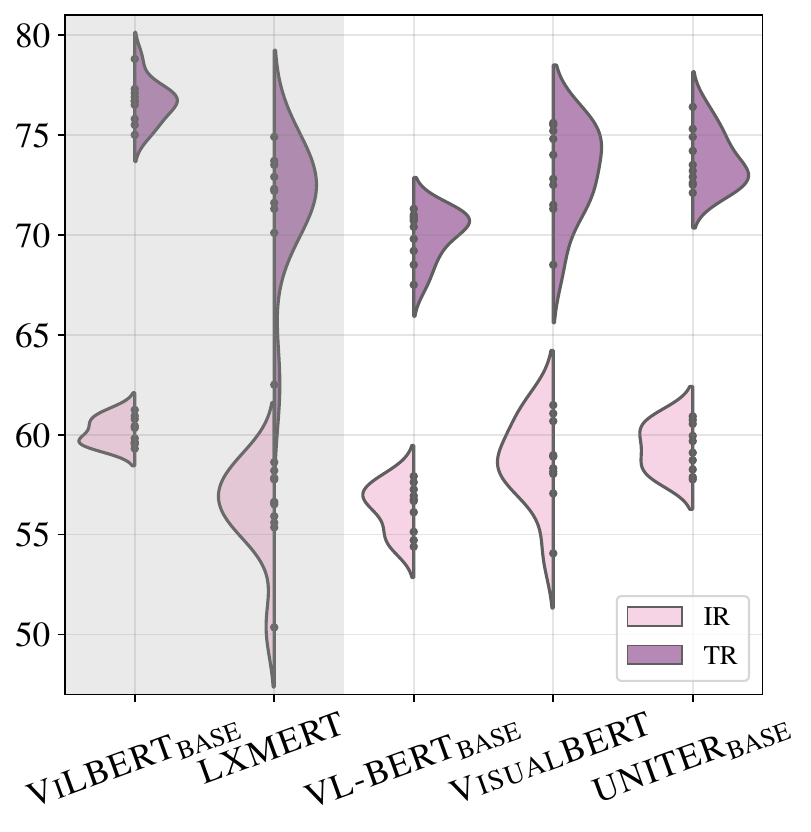}
  \caption{Flickr30k}
  \label{fig:retrieval_pre}
\end{subfigure}
\caption{Pretraining variance of \vl \berts. Each model is pretrained $10$ times and fine-tuned once.}
\label{fig:pretrain}
\end{figure*}

\begin{figure}[t]
\centering
\begin{subfigure}{.245\textwidth}
  \centering
  \includegraphics[width=\linewidth]{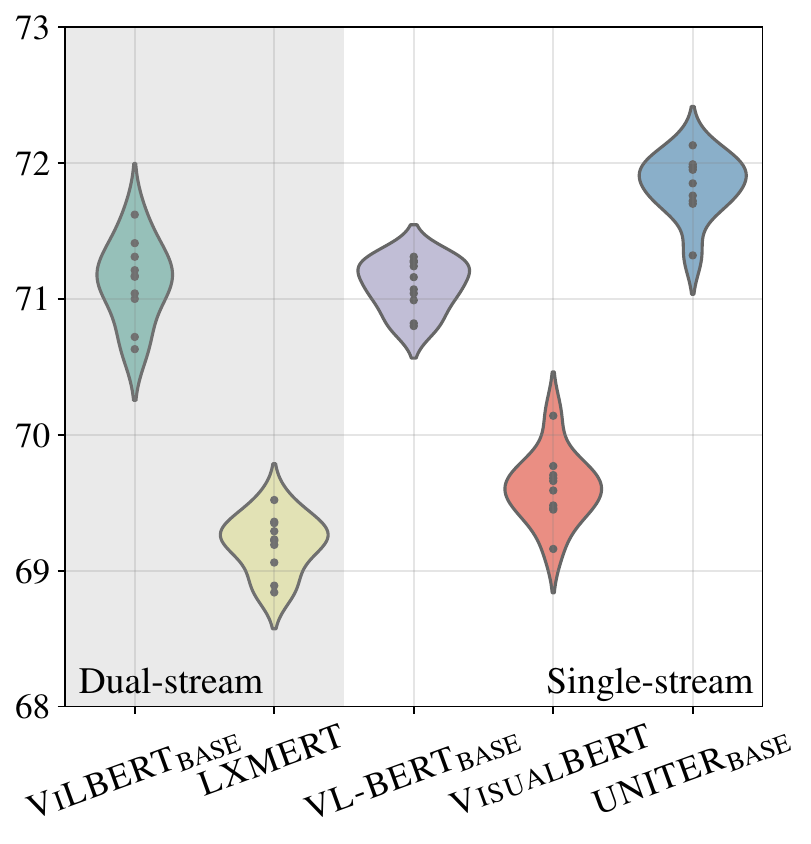}
  \caption{RefCOCO+}
  \label{fig:refcoco_fine}
\end{subfigure}%
\begin{subfigure}{.245\textwidth}
  \centering
  \includegraphics[width=\linewidth]{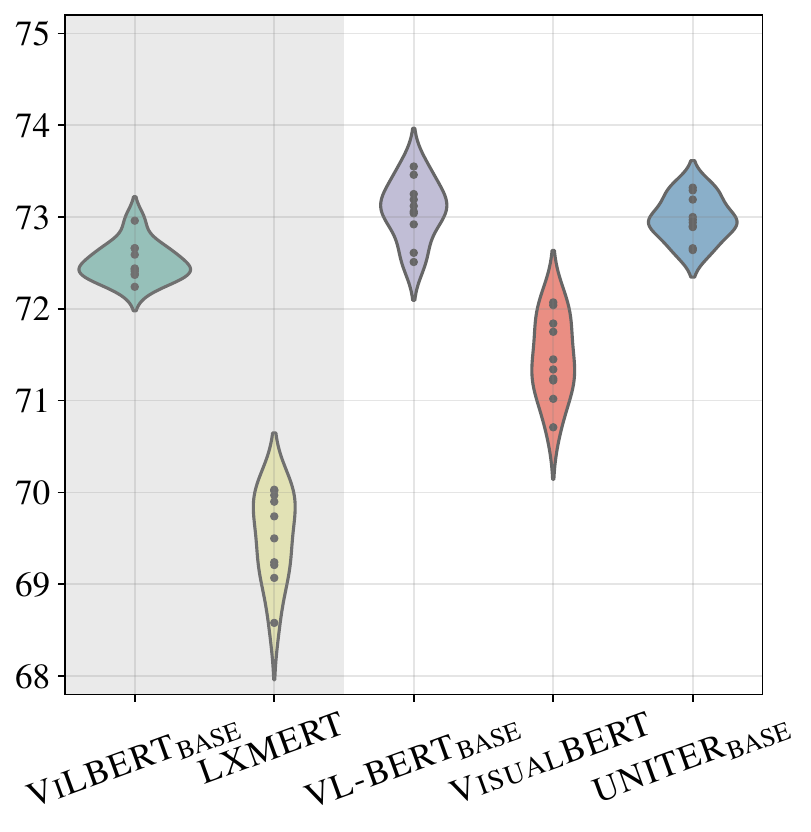}
  \caption{NLVR2}
  \label{fig:nlvr_fine}
\end{subfigure}
\caption{Fine-tuning variance of \vl \berts on RefCOCO+ and NLVR2. Each model is pretrained once and fine-tuned $10$ times on each task.}
\label{fig:fine-tune}
\end{figure}

\begin{figure}[t]
\centering
\begin{subfigure}{.245\textwidth}
  \centering
  \includegraphics[width=\linewidth]{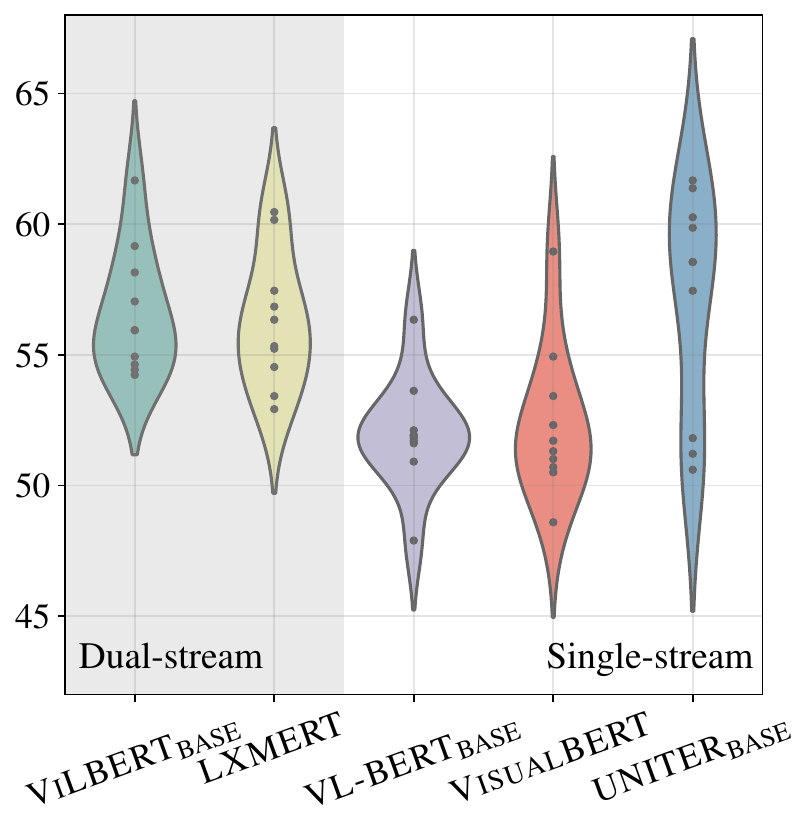}
  \caption{Pretraining}
  \label{fig:contrast_pre}
\end{subfigure}%
\begin{subfigure}{.245\textwidth}
  \centering
  \includegraphics[width=\linewidth]{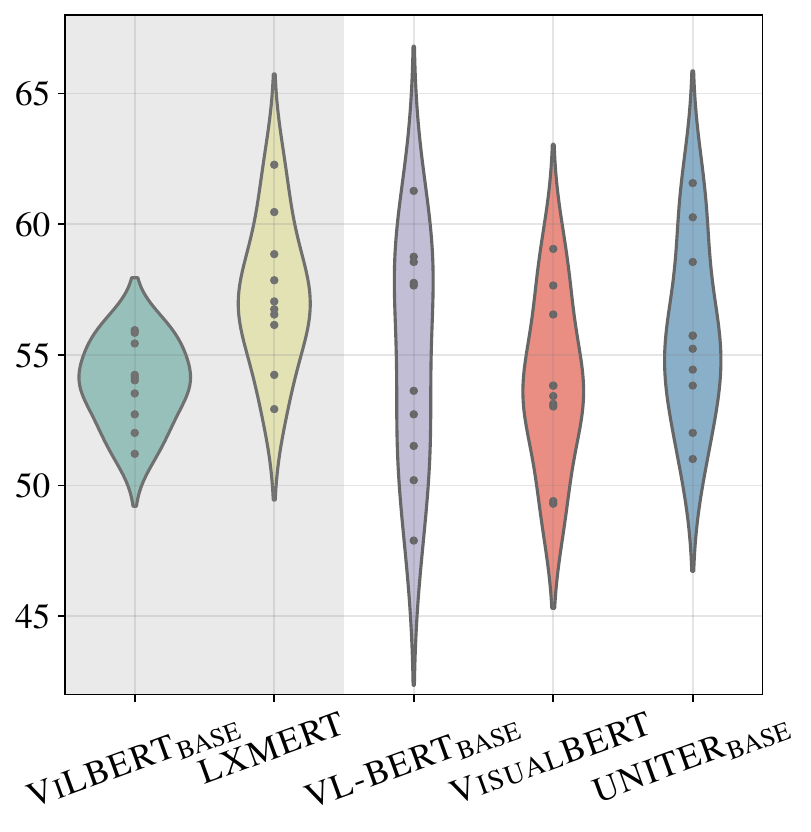}
  \caption{Fine-tuning}
  \label{fig:contrast_fine}
\end{subfigure}
\caption{Variance of \vl \berts on the Constrastive Set of NLVR2, when each model is pretrained $10$ times and fine-tuned once (a), or pretrained once and fine-tuned $10$ times (b).}
\label{fig:contrast}
\end{figure}

\paragraph{Results}
\cref{tab:controlled} shows the results of our controlled study.
First, we note that the performance of \vilbert and \vlbert is similar compared to training with their original hyperparameters.
In fact, VQAv2 performance improves for \vilbert, showing that dual-stream models do \emph{not} require different sizes in the two streams.
\vlbert also performs similarly to its official setup, showing that the additional ITM pretraining objective in our controlled setup does not hurt downstream task performance (contrarily to the results reported in their paper).
We do, however, note that \lxmert performs worse on NLVR2 and VQAv2 in our controlled setup than with its original hyperparameters, suggesting that \lxmert may require more pretraining steps to converge.
Overall, the results show that most of the examined models perform similarly in our controlled setup, compared to the official setups.

\subsection{Fine-tuning Variance}

We now turn our attention to the effect of fine-tuning variance on task performance. 
It has been observed that the fine-tuning of \bert is sensitive to randomness in initialisation and data ordering~\citep{dodge2020fine}. 
Here, we investigate the sensitivity of the five models used in the controlled study. 
We fine-tune each model $10$ times on the RefCOCO+ and NLVR2 tasks by varying the seed. This changes training data order and the weight initialisation of the classification layer.
\cref{fig:fine-tune} shows violin plots of the distribution of results, in which the dots represent the experimental observations.
We also report an average standard deviation of $0.3$ points for these models across both tasks.
However, the minimum and the maximum scores of a given model often differ by $1$ or more points, showing how \emph{a single fine-tuning} run of these models can lead to \emph{incorrect} conclusions.

\subsection{Pretraining Variance}

\begin{figure*}
\centering
\begin{subfigure}{.15\textwidth}
  \centering
  \raisebox{10mm}{\includegraphics[width=\linewidth, trim={0cm 0cm 14cm 0cm}, clip]{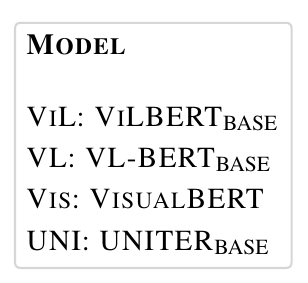}}
\end{subfigure}%
\begin{subfigure}{.185\textwidth}
  \centering
  \includegraphics[width=\linewidth]{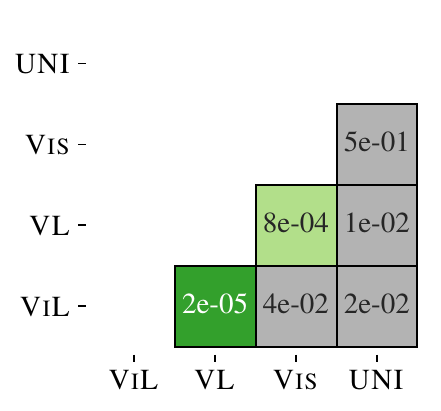}
  \caption{VQAv2}
  \label{fig:vqa_perm}
\end{subfigure}%
\begin{subfigure}{.15\textwidth}
  \centering
  \includegraphics[width=\linewidth]{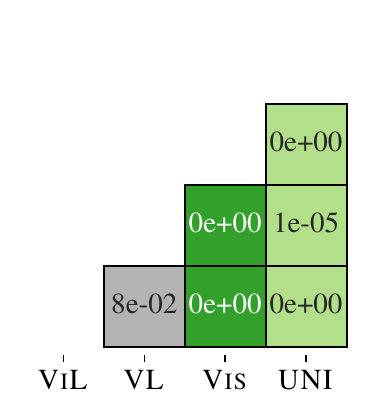}
  \caption{RefCOCO+}
  \label{fig:ref_perm}
\end{subfigure}%
\begin{subfigure}{.15\textwidth}
  \centering
  \includegraphics[width=\linewidth]{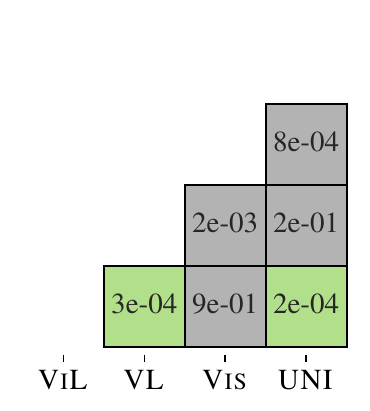}
  \caption{NLVR2}
  \label{fig:nlvr_perm}
\end{subfigure}
\begin{subfigure}{.15\textwidth}
  \centering
  \includegraphics[width=\linewidth]{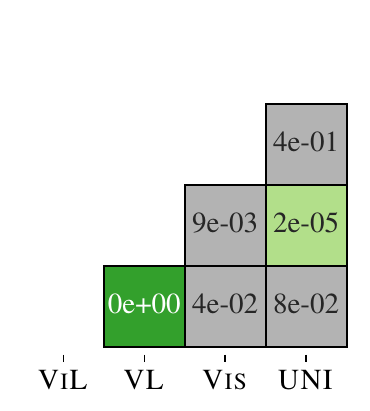}
  \caption{Flickr30k IR}
  \label{fig:ir_perm}
\end{subfigure}%
\begin{subfigure}{.15\textwidth}
  \centering
  \includegraphics[width=\linewidth]{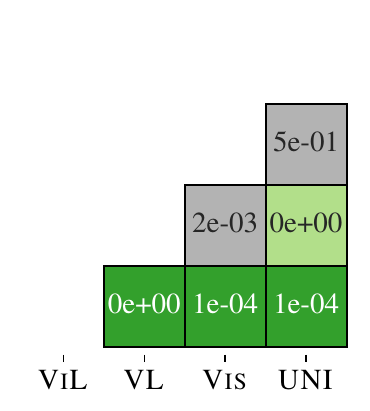}
  \caption{Flickr30k TR}
  \label{fig:sub2}
\end{subfigure}
\caption{Exact test between any two \vl \berts. Each box shows the p-value for the corresponding pair of models. Green boxes denote statistical significance at $0.005$ after Bonferroni correction. Boxes are dark green if the model in the y-axis outperforms the one in the x-axis, and vice versa for light green.}
\label{fig:perm}
\end{figure*}

In the previous section, we found substantial variance in the performance of \vl \berts across $10$ fine-tuning runs. 
We now investigate if the pretraining phase is similarly affected by different runs. 
Here, each model in our controlled setup is pretrained $10$ times and fine-tuned once on four tasks: VQAv2, RefCOCO+, NLVR2, and Flickr30K image--text retrieval. 
By varying the seed, we modify training data order as well as all the layers that are not initialised from BERT (e.g., the visual embeddings, the masked object classification head and the ITM head in single-stream models).
\cref{fig:pretrain} shows violin plots for each task.
We start by noting that our first pretraining run (\cref{tab:controlled}) of \lxmert was the worst one (its text retrieval recall on Flickr30K is $10$ points lower than its mean). 
We also confirm that \lxmert has slower convergence rate, with its task performance after $10$ epochs showing the largest variance among the \vl \berts we tested.
On the other hand, we find that some of these architectures are less prone to variance caused by pretraining seed, such as \vilbert for VQA and retrieval tasks, and \uniter for referring expression.
Nevertheless, the performance of all of these models can vary by more than $1$ point in several tasks solely due to random initialisation.

\subsection{Evaluating Local Decision Boundaries}
Previous work has shown that state-of-the-art systems can exploit systematic gaps in the data to learn simple decision rules that let them achieve high performance on test data~\cite{gururangan-etal-2018-annotation,geva-etal-2019-modeling,ribeiro-etal-2019-red}.
In an effort to more accurately estimate model performance, \citet{gardner-etal-2020-evaluating} proposed \emph{contrast sets}: datasets in which existing test instances have small but label-changing modifications in order to characterise the correct decision boundary near them.
\cref{fig:contrast} shows the performance of our analysed models on the NLVR2 contrast set.
Similar to \citet{gardner-etal-2020-evaluating}, we see that \lxmert loses around $15$ points when evaluated on perturbed samples.
Furthermore, models that performed much better on the standard test set now achieve comparable performance to \lxmert, showing that they exploited systematic gaps.
That is, all of these \vl \berts would perform similarly when evaluated on out-of-distribution data.

\subsection{Single- or Dual-stream Architectures}

\begin{table}
\setlength\tabcolsep{5pt}
\footnotesize
\center
  \begin{tabular}{lrrrr}
  \toprule
  \multirow{2}{*}{\small \shortstack{\textbf{Dataset}}} & \multicolumn{2}{c}{\textbf{Single/Dual Stream}} & \multicolumn{2}{c}{\textbf{\vl \berts}} \\
  \cmidrule(r){2-3}
  \cmidrule(r){4-5}
   & \multicolumn{1}{c}{{F-test}} & {p-value} & \multicolumn{1}{c}{{F-test}} & {p-value} \\
  \midrule
  \band VQAv2 & 11.40 & 1.7e-03 & 12.75 & 8.0e-06$^{*}$ \\
  RefCOCO+ & 0.10 & 7.6e-01 & 111.61	& 2.7e-18$^{*}$ \\
  \band NLVR2 & 8.28 & 6.5e-03 & 13.41	& 5.0e-06$^{*}$ \\
  Flickr30k IR & 9.64 & 3.6e-03 & 13.27 & 5.0e-06$^{*}$ \\
  \band Flickr30k TR & 31.14 & 2.0e-06$^{*}$ & 29.74 & 7.5e-10$^{*}$ \\
  \bottomrule
  \end{tabular}
  \caption{ANOVA between single- and dual-stream architectures (left) and between all the tested \vl \berts (right). $^{*}$ denotes significant results at $p<0.005$ after Bonferroni correction.}\label{tab:stream_anova}
\end{table}

One of the key design choices that distinguishes \vl \berts is the number of ``streams'' used by the encoder to process visual and linguistic inputs.
\citet{lu2019vilbert} showed how their single-stream baseline performed worse than their dual-stream \vilbert architecture, while \citet{chen2020uniter} claimed single-stream \uniter outperformed \vilbert.
Our controlled study across several tasks and different pretraining initialisations allows us to provide an answer grounded with statistical tests.
To do so, we split the models in dual- and single-stream architectures\footnote{We only consider \vilbert for dual-stream encoders due to \lxmert's sub-optimal performance in our setup.} and run a \emph{one-way ANOVA} (\cref{tab:stream_anova}). After Bonferroni correction, we only find statistical difference at $p<0.005$~\cite{benjamin2018redefine} between these two groups for the Flickr30K text retrieval task.

On the other hand, running the same test among the various \vl \berts, without grouping them as single- or dual-stream architectures, returns statistical significance in each task (\cref{tab:stream_anova}).
This table tells us that the null hypothesis, the models have the same average performance, does not hold.
However, it does not allow us to discern where statistical differences lie.
To do so, we conduct a post-hoc \emph{exact test} at significance level $p<0.005$.
\cref{fig:perm} shows the corresponding pair-wise p-values and highlights significant differences between any two models after Bonferroni correction.
For instance, \vilbert is significantly different compared to all other models in text retrieval on Flickr30k, while \uniter is significantly different on RefCOCO+.

\subsection{The Importance of the Embeddings}

\begin{figure}[t]
\begin{center}
  \includegraphics[width=\linewidth, trim={0cm 0.2cm 0cm 0cm}, clip]{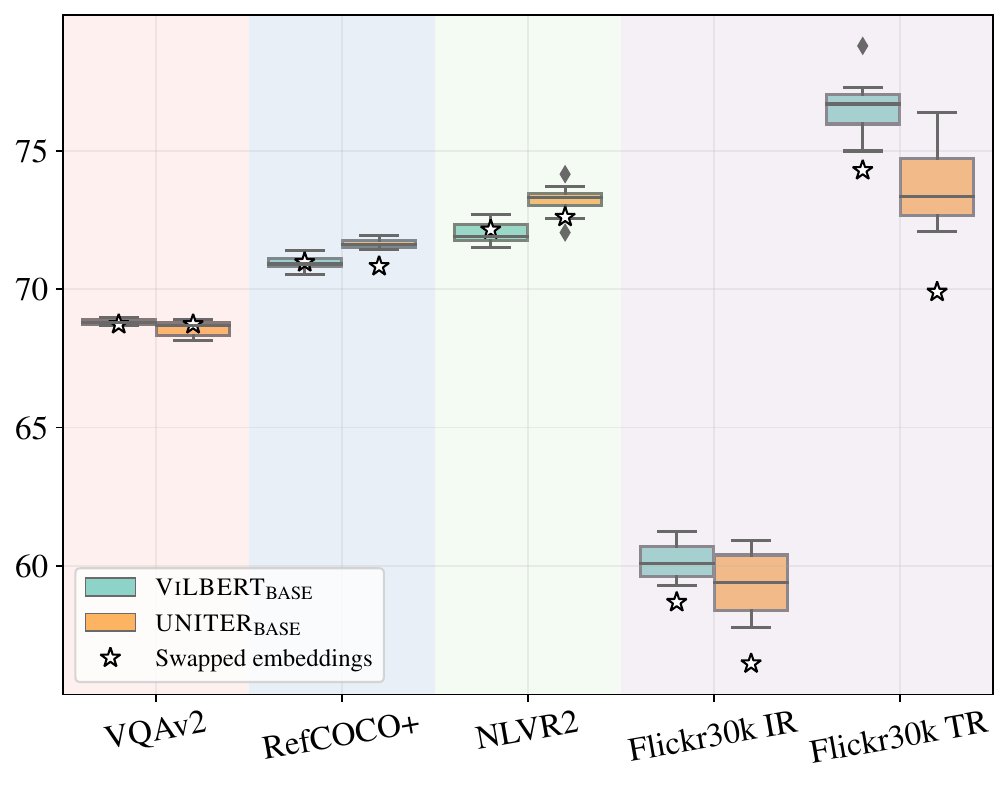}
\end{center}
  \caption{Results of swapping \vilbert and \uniter embeddings ($\star$) compared to their performance when pretrained $10$ times (box plots).} \label{fig:swap}
\end{figure}

Finally, our controlled setup leads us to an interesting finding: the embedding layer ({\cref{sec:embs}) plays a crucial role in the final performance of \vl \berts.
In fact, the only difference among \vlbert, \visualbert and \uniter in our setup is their embedding layer.
\cref{fig:pretrain} and \cref{fig:fine-tune} show that this can have a drastic impact on the downstream performance although the literature has given little attention to this detail.
For instance, \citet{chen2020uniter} claim that the main contribution of \uniter is the set of pretraining tasks, while our results, wherein all the models are trained on the same pretraining tasks, highlight that their embedding layer is an important confound on final performance.
Interestingly, \visualbert is the only model that \emph{does not} encode the locations of regions of interest in its embeddings. 
This leads it to considerably lower performance on RefCOCO+, showing that this information is extremely useful for this task.

Given this result, we conduct one additional experiment to see whether the embedding layer biased our conclusion for dual- and single-stream performance.
To test this, we swap the embedding layers of \vilbert (best dual-stream) and \uniter (overall better single-stream) with each other, which we pretrain and fine-tune once (\cref{fig:swap}).
Similar to our previous results, embeddings are especially important for the tasks of referring expression and retrieval. 
However, no single embedding layer performs better, corroborating that dual- and single-stream architectures perform on par and showing that different embedding strategies are necessary to maximise performance in these two families of \vl \berts.

\subsection{Limitations}
All the experiments in this paper are limited to models that use a specific type of pretrained and frozen visual encoder.
While most \vl \berts follow this paradigm, some studies find beneficial to jointly learn the visual encoder with language~\cite{Su2020VL-BERT:,huang2020pixel,radford2021learning,kim2021vilt}.
In addition, we only consider base architecture variants (initialised with \bert{}$_{\textit{BASE}}$) and pretrained on CC.
Studying the effects of visual encoders, pretraining data and larger models is left as future work.

Although we expect longer pretraining would be beneficial for every model, in our controlled setup, we pretrain each model for $10$ epochs to reduce resource consumption.
Here, we also constrain our hyperparameter search over a small grid of values that have been used in the literature.
Finally, we leave a thorough, controlled study of the various pretraining objectives to future work.

\section{Reproducibility and the Environment} \label{sec:discussion}

From the perspective of reproducible research, there are several advantages to using the \volta framework for \vl encoders.
First, \volta reduces confounds due to differences in implementations, while also enabling fair comparisons with related work.
Second, visual and textual data only need to be preprocessed \emph{once} instead of creating model-specific formats for every \vl \bert.

From a financial perspective, the costs involved in pretraining hampers contributions from many academic institutions and deters the evaluation of multiple trained models, which we showed to be extremely important for \vl \berts.
We estimate that pretraining a single model $10\times$ in our controlled setup for $4$ downstream tasks requires a $4$-GPU machine on AWS for two months, at a cost of $\sim\$6{,}000$, corresponding to $200$ GPU-compute days. 
Fortunately, we had access to an internal server, but the our experiments still required $1{,}500$ GPU days for training and evaluation. 
While we were able to reduce the financial costs, there are severe environmental and carbon footprint costs in \vl pretraining~\citep{strubell-etal-2019-energy}.\footnote{We distribute many of our pretrained \vl \berts in \volta to amortise the environmental costs.} 

We hope that \volta will serve as a basis for research in \vl pretraining, enabling easy and fair comparisons across architectures, and ensuring that progress is not obfuscated by confounds.
\section{Conclusion}
We introduced and implemented a unified mathematical framework, under which recently proposed \vl \berts can be specified as special cases.
We conducted a series of controlled studies within this framework to better understand the differences between several models.
We found that the performance of the considered models varies significantly due to random initialisation, in both pretraining and fine-tuning. 
We also found that these models achieve similar performance when trained with the same hyperparameters and data.
Notably, some models outperform others but we found that (a) single- and dual-stream model families are on par, and (b) embedding layers play a crucial role towards a model's final performance.

Our fast-paced field rewards the contribution of new methods and state-of-the-art results \cite{rogers-augenstein-2020-improve}, which often contrasts with controlled comparisons and training multiple models for variance estimation. In this paper, we showed that several methods for vision-and-language representation learning do not significantly differ when compared in a controlled setting. This finding echoes similar studies of variants of LSTMs \cite{greff2016lstm} and Transformers \cite{narang2021transformer} that are not significantly better than the original models. Looking to the future, we recommend that new \vl \berts are pretrained on similar datasets, and that researchers report fine-tuning variance, in addition to their best performing model. We hope that our findings will encourage more controlled evaluations of newly proposed architectures for vision-and-language and beyond.

\section*{Acknowledgments}
{\scriptsize\euflag} We are grateful to the action editor Jacob Eisenstein and the anonymous reviewers at TACL for their constructive comments and discussions. This project has received funding from the European Union's Horizon 2020 research and innovation programme under the Marie Sk\l{}odowska-Curie grant agreement No 801199 and by ``Research and Development of Deep Learning Technology for Advanced Multilingual Speech Translation,'' the Commissioned Research of National Institute of Information and Communications Technology (NICT), Japan.

\bibliography{references}
\bibliographystyle{acl_natbib}

\end{document}